\documentclass[11pt,twocolumn]{IEEEtran}
\usepackage{amsmath}
\usepackage{amsfonts}
\usepackage{amssymb}
\usepackage{graphicx}
\usepackage{subfigure} 

\begin{document}
\title{Efficient Image Retargeting for High Dynamic Range Scenes} 
\author{Govind Salvi, Puneet Sharma, and Shanmuganathan Raman} 
\date{ }
\maketitle
\begin{abstract}
Most of the real world scenes have a very high dynamic range (HDR). The mobile phone cameras and the digital cameras available in markets are limited in their capability in both the range and spatial resolution. Same argument can be posed about the limited dynamic range display devices which also differ in the spatial resolution and aspect ratios.

In this paper, we address the problem of displaying the high contrast low dynamic range (LDR) image of a HDR scene in a display device which has different spatial resolution compared to that of the capturing digital camera. The optimal solution proposed in this work can be employed with any camera which has the ability to shoot multiple differently exposed images of a scene. Further, the proposed solutions provide the flexibility in the depiction of entire contrast of the HDR scene as a LDR image with an user specified spatial resolution. This task is achieved through an optimized content aware retargeting framework which preserves salient features along with the 
algorithm to combine multi-exposure images. We show the proposed approach performs exceedingly well in the generation of high contrast LDR image of varying spatial resolution compared to an alternate approach.
\end{abstract}

\section{Introduction}
\label{sec:1}
\begin{figure}[!htbp]

\centering
\subfigure{
\includegraphics[width=.33\columnwidth]{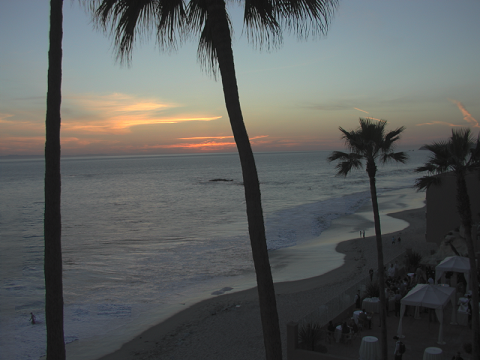}
\includegraphics[width=.33\columnwidth]{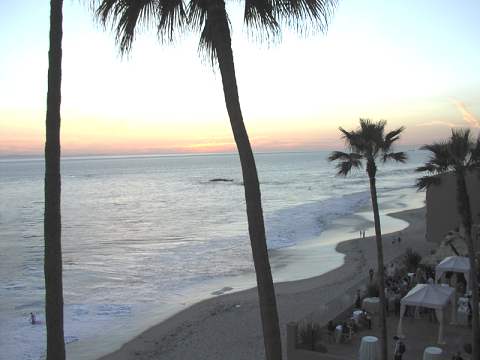}
\includegraphics[width=.33\columnwidth]{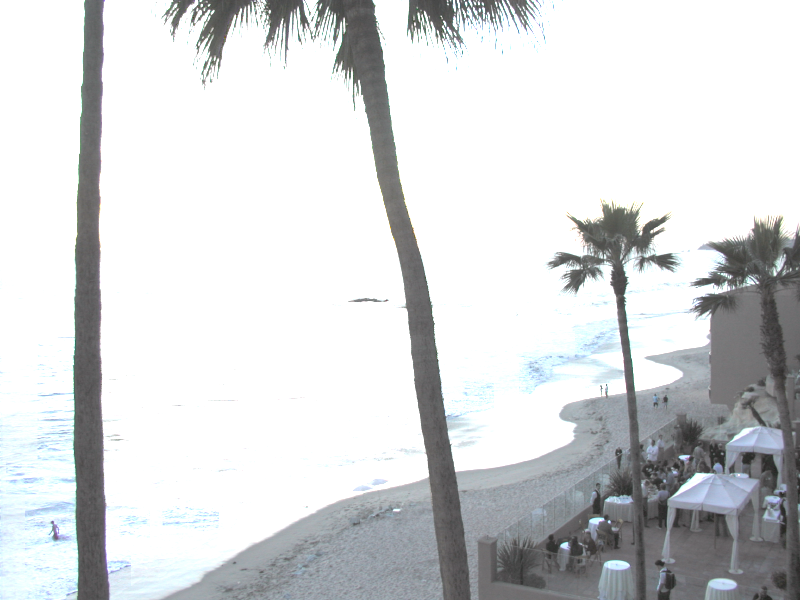}}

\subfigure{
\includegraphics[width=.33\columnwidth]{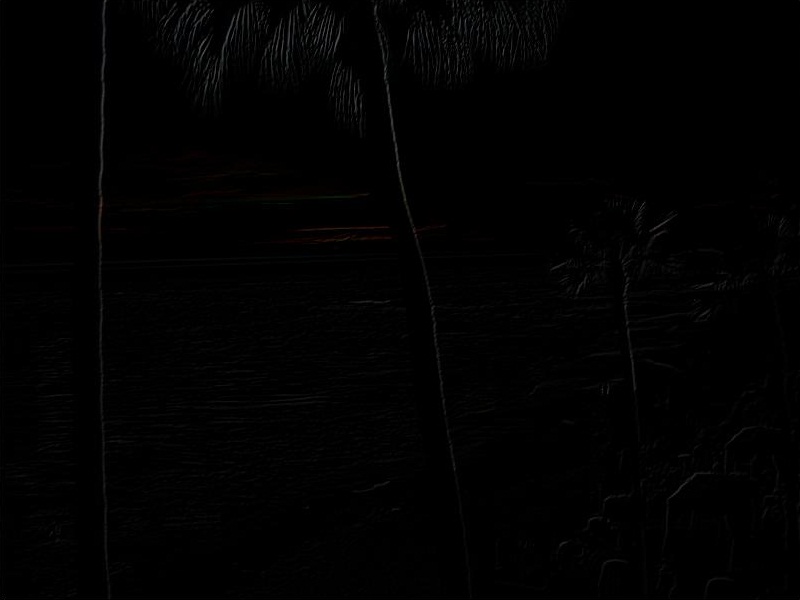}
\includegraphics[width=.33\columnwidth]{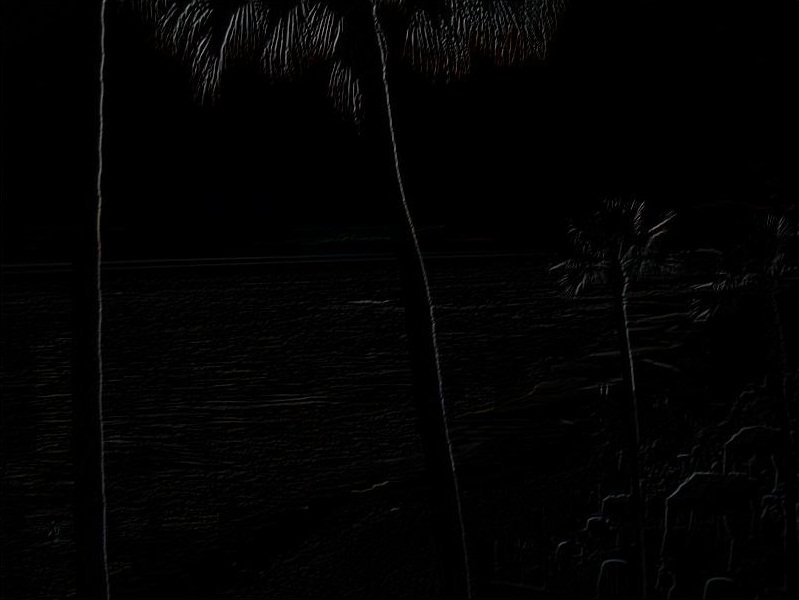}
\includegraphics[width=.33\columnwidth]{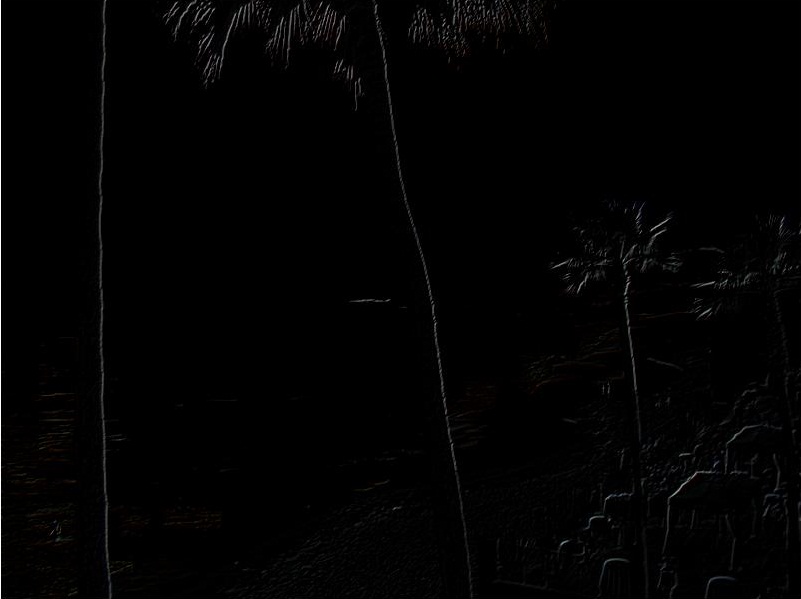}}

\subfigure{
\includegraphics[width=.33\columnwidth]{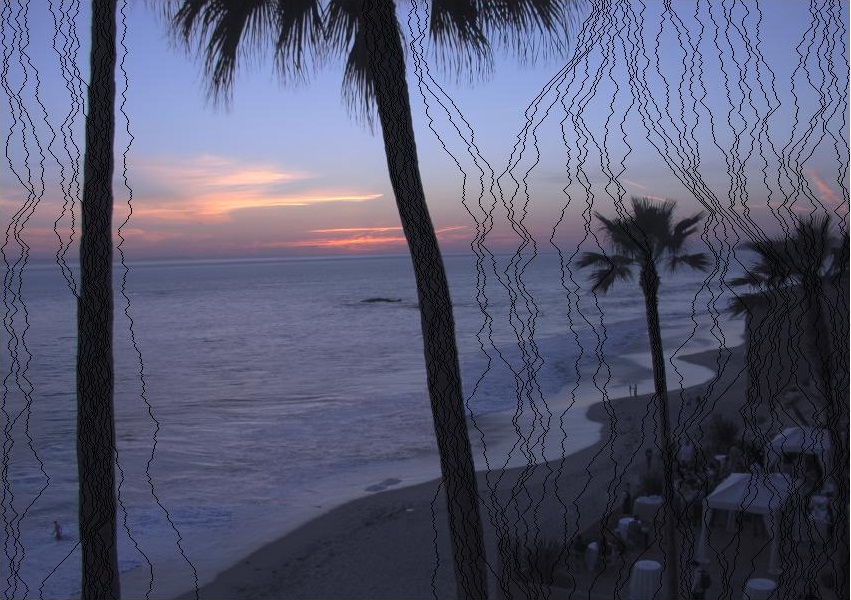}
\includegraphics[width=.33\columnwidth]{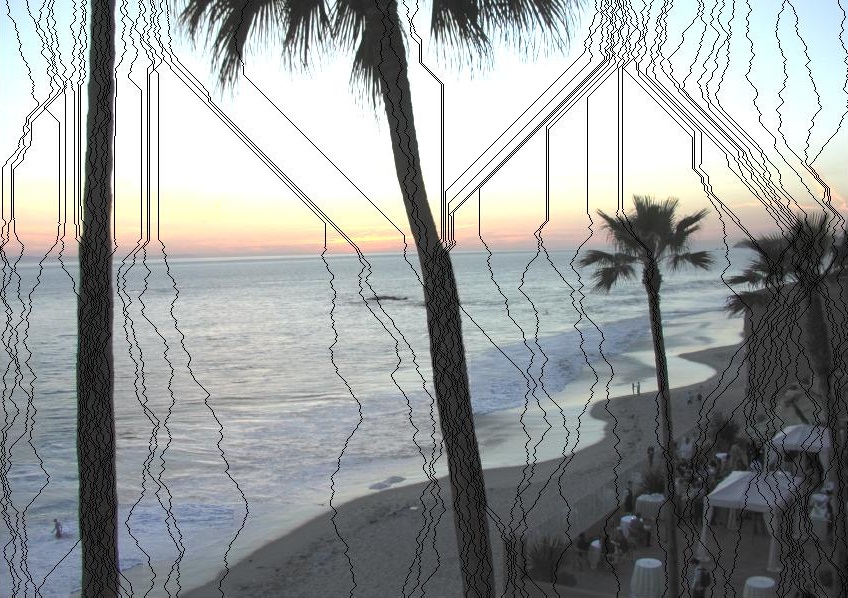}
\includegraphics[width=.33\columnwidth]{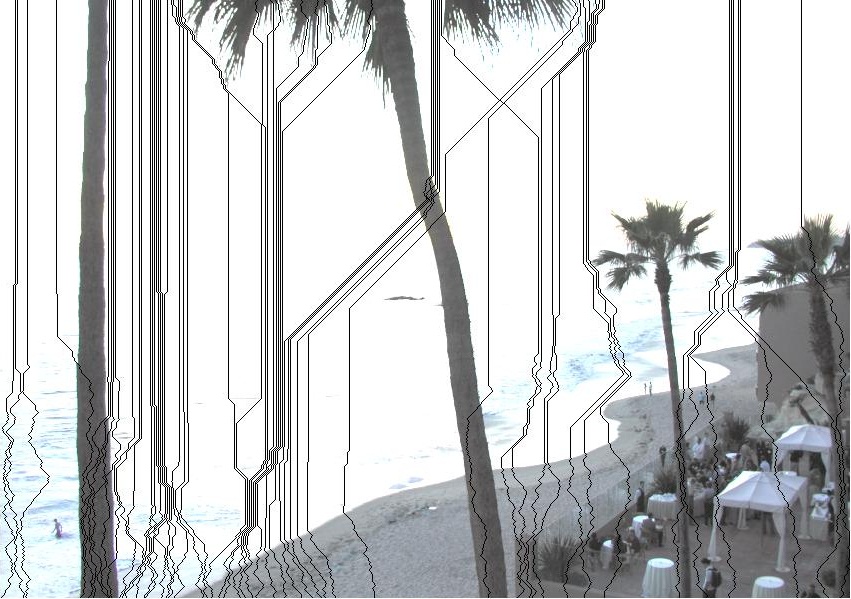}}

\caption{Top Row: Bracketed LDR exposure sequence, Middle Row: Energy function corresponding to each LDR image, Last Row: vertical seams found by the algorithm. Notice that on each image the minimum energy seams found by the algorithm are different.}
\label{beach}
\end{figure}
Real world scenes have a high dynamic range (HDR). An example of such a HDR scene is one which has both brightly and poorly lit regions. This implies that the range of brightness levels are very high. Human visual system (HVS) can visualize all the brightness levels of the scene through visual adaptation. Even analog cameras can capture major percentage of the brightness levels. The digital capturing devices such as mobile phone cameras and digital cameras can not capture the entire HDR of a given scene. The digital cameras are  limited in terms of their spatial resolution as evident by the spatial resolution in various digital imaging sensor architectures. In other words, digital cameras have limited range and spatial resolutions which are caused primarily due to the limitations posed by the imaging sensor design. It is highly complex to capture all the brightness levels of a HDR scene in finite duration.

Limited dynamic range is caused mainly due to the limited well capacity of the sensor elements. The dynamic range of the image can be enhanced by the HDR imaging techniques which rely on the capture of multi-exposure low dynamic range (LDR) images of the scene \cite{Reinhard10}. These approaches recover the camera response function (CRF) of the imaging system and employ it to create the HDR image of the scene. The generated HDR images are then tone mapped into a high contrast LDR image compatible with a given digital display device. Alternately, the high contrast LDR image of the scene can be directly generated without the knowledge of CRF by-passing the HDR imaging pipeline.

The spatial resolution of the image can either be reduced or enhanced by employing super-resolution algorithms \cite{Park03}. These techniques perform resolution change through efficient interpolation without preserving the salient contents of the scene. The image retargeting approaches which have recently been developed enable one to change the spatial resolution of the image while preserving the contents of the image which are important \cite{Avidan07}. Retargeting has been the standard approach when one wants to modify the spatial resolution of a given image.

Consider a set of multi-exposure images of a scene captured using traditional technique such as Auto Exposure Bracketing (AEB). The problem we would like to address is whether we can achieve the flexibility in both the spatial and range resolutions given a set of multi-exposure images corresponding to a static scene. The obvious solution to this problem is to first generate a HDR image using standard approach and then perform spatial resizing either by super resolution or by image retargeting. The question to be answered while using such a solution is this: whether this approach is the optimal one, or can we derive a better optimal solution. This work is primarily focused on exploring alternate better solutions to this challenging problem.

The main objective of this work is to search for an optimal solution to achieve a flexible range (contrast) and spatial resolution, given a set of muti-exposure LDR images of a static scene. We develop an algorithm to achieve such an optimal solution to this problem. We show that the proposed approach performs far better than the obvious solution and leads to the generation of a high contrast LDR image with provision to adapt the size of the image compatible with a given display device. The key contributions of this novel approach are the algorithms which achieve the following tasks.
\begin{itemize}
 \item Flexible content aware spatial retargeting of an image corresponding to a static HDR scene,
 \item Depiction of high contrast information within the user specified spatial resolution,  
 \item Achieving high quality desired LDR images without any visible artifacts, and
 \item Assumption: No knowledge of exposure times, scene information, and CRF. 
\end{itemize}

The paper is organized as follows. We shall review the prior relevant work in section~\ref{pre} which are key to our discussions later on. We present the primary motivation behind the present work in section~\ref{motivation}. We shall discuss the proposed algorithm for simultaneous contrast and content-aware spatial retargeting  in detail in section~\ref{algo}. Section~\ref{result} presents the results corresponding to various aspects of the proposed solution. We conclude the paper in section~\ref{con} summarizing the key contributions and presenting some pointers on future enhancement of the proposed approach.

\section{Previous Work}
\label{pre}

In recent times, creation of images which depict all the brightness levels in a natural scene has been a topic of great interest. Various research groups have been working on this topic and have proposed various solutions to this challenging problem \cite{Reinhard10}. A bracketed exposure sequence, which spans the entire dynamic range of the real world scene, comprises of a set of LDR images that are shot with a digital camera. The CRF should be recovered in order to linearize the intensities. The HDR image can be generated by compositing these multi-exposure images in linearized intensity domain  (\cite{Mann95}, \cite{Debevec97}). The HDR images can be displayed in specialized HDR displays \cite{Seetzen04}. However, for visualizing the generated HDR image in common LDR displays we need to perform tone reproduction operation. Many different tone mapping operators have been proposed in recent years with various performance levels for different scenes (\cite{Reinhard02}).

On the other hand, exposure fusion approaches relieve us the need of intermediate HDR image generation and tone mapping operation (\cite{Mertens07}, \cite{Raman07}). Exposure fusion involves compositing the different Laplacian pyramid levels of the multi-exposure images with appropriate weights in order to reduce saturation and enhance contrast (\cite{Mertens07},\cite{Burt83}). Similar approach can be further used for merging flash/no-flash images to get the best information out of both the images and create a better image (\cite{Petschnigg04}, \cite{Eisemann04}). Dynamic scenes captured with the help of multi-exposure images lead to artifacts which requires appropriate deghosting prior to compositing \cite{Khan06}.  Recently, researchers have turned their attention to reconstruct a HDR image of a non-static scene  with the knowledge of CRF \cite{Gallo09} and without the knowledge of CRF (\cite{Pece10}, \cite{Raman11}, \cite{Zhang12}). The generation of a HDR image from a set of multi-exposure images when 
both the camera and scene change has been addressed in the recent works (\cite{Zimmer11}, \cite{Hu12}, \cite{Sen12}).

Image resizing is a different problem in which one attempts to change the spatial resolution of the given image popularly known as image super-resolution. Super-resolution can be achieved by using multiple images of the same scene with sub-pixel shifts \cite{Park03}. Content aware resizing should be done in a way that minimizes the amount of important information we lose during resizing operation. Approaches such as face detectors and visual saliency map detectors can be used to achieve this task (\cite{Viola01}, \cite{Itti98}). After creating a visual saliency map, image can be cropped to capture the most salient regions in the image. These methods are based on the conventional technique of either cropping or removal of columns and rows.

These methods are often constrained by the ratio to which a given image can be resized. Resizing the image beyond a critical factor generates a high degree of artifacts. Recently, methods have been proposed by which this critical ratio can be improved. Changing the spatial resolution of the image is also important and used enormously in texture synthesis, the goal here is to generate a large textured image from a small textured image \cite{Efros01}. But the solution in texture synthesis can not be extended to natural scenes directly as they follow complex statistics. A natural image may have multiple different regions of importance and sometimes a user interaction is exploited to specify the regions which are of greater importance \cite{Agarwala04}.

Image retargeting is a much better automatic approach which has been widely used for content aware resizing \cite{Avidan07}. The first popular implementation of image retargeting, seam carving, involves the identification of minimum energy seams which have to be removed or added so that there is minimum loss of information. An efficient energy metric based on gradient measure serves as the energy function. Optimal seam carving can alternately use different types of energy functions such as gradient magnitude, entropy, visual saliency, eye-gaze movement, and more. The removal or insertion of seams can be done in such a way as to make it compatible with the resolution and aspect ratio of the display device. Seam carving can be extended to perform video retargeting (\cite{Rubinstein08}, \cite{Rubinstein10}). An overview of the different types of image retargeting approaches can be found in the recent tutorial \cite{Banterle11}. 

There is always a trade-off between the spatial resolution and the range resolution of an imaging sensor. A typical example is the assorted pixels which use multiple sensor elements with different sensitivities to create a HDR image \cite{Narasimhan05}. Here, we sacrifice some spatial resolution to gain more dynamic range. The size of the sensor element can not be made smaller than a particular size due to noise and limited well capacity (\cite{Gamal02}, \cite{Granados10}, \cite{Hasinoff10}). These studies on the imaging sensor emphasize the need for creating a new application with flexible spatial and range resolutions.

\section{Motivation}
\label{motivation}
The recent work for spatial as well as dynamic range improvement from a set of multi-exposure images requires one to capture multi-exposure images with subpixel shifts \cite{Zimmer11}. This approach is a combination of the traditional HDR imaging and super-resolution approaches posed in a unified optimization framework. Therefore, this method does not enable one to perform content aware resizing though it helps in improving the dynamic range and the spatial resolution. Existing methods on simultaneous improvement of spatial resolution and dynamic range do not take into considertion, the content present in the image (\cite{Gunturk06}, \cite{Choi09}, \cite{Zimmer11}).

The primary motivation behind this work is to generate a high contrast LDR image corresponding to a given HDR scene with flexible content aware image resizing capability. This application is quite useful in the present scenario as we have digital display devices which have different spatial resolution and aspect ratios but can only display LDR content. Examples of such display devices include Apple iPad, smartphones and tablets by Nokia and Samsung, netbooks, etc. The trivial solution to this problem as discussed eariler is to fuse the multi-exposure images and then to retarget the resultant image spatially in order to make it compatible with a given display device. This work is an attempt to probe for alternate efficient solutions for this problem and show how such solutions can indeed be better than the trivial solution in terms of image contrast and lesser artifacts incurred. 

The main objective behind this work is to find an efficient way to merge multiple differently exposed images of a static scene into a high contrast LDR image with flexible spatial resolution. This task is achieved by an efficient algorithm which performs this task while reducing the loss in contrast and reducing any artifacts in the final LDR image. We shall present the basic algorithm behind the proposed approach in the next section.

\section{Proposed Algorithm}
\label{algo}
In this section we propose multiple approaches for efficient retargeting of a HDR scene. Our algorithm uses a set of LDR images having different exposure times. The input images are registered LDR images of the same static scene. Let $I_{1}, I_{2}, I_{3},..,I_{N} $ be the set of input LDR images. We use magnitude of the gradient as the energy metric. One can use other energy metrics like entropy, visual saliency, also \cite{Avidan07}.

\begin{equation}
\hskip .25\columnwidth E=\left| {\partial I\over\partial x}\right| +\left| {\partial I\over\partial y}\right| 
\end{equation}

Through this energy metric we generate a cumulative energy metric, which enables us to find the minimum energy seams (seams with least importance) in individual LDR images. We shall start the discussion with the trivial approach for retargeting LDR image corresponding to a HDR scene. We assume that we do not know the exposure times and the CRF in the present work.

\subsection{Direct Approach}
One of the approaches for resizing image corresponding to a HDR scene is to take multiple LDR images of the scene with different exposure times and subjecting them to exposure fusion \cite{Mertens07}. This approach results in an image having much higher contrast than the individual input images. Further applying optimal seam carving on this high contrast image yields the resized high contrast LDR image of the scene.

This method is constrained by the ratio upto which a certain image can be resized. Increasing or decreasing the aspect ratio of an image beyond a critical factor can produce artifacts of greater magnitude (see figure~\ref{chameleon}(b) and figure~\ref{beach1}(a)). Suppose we have multiple images of the same scene with different exposure times, we obtain multiple seams with least energy for each image in the given LDR image set. We shall show how removing or adding seams with minimum energy (before applying exposure fusion) and then using exposure fusion yields to a better quality high contrast LDR image.

\subsection{Statistical Approach} 
In this approach, for a given energy metric we first find the cumulative energy matrix for individual LDR image. Consequently, with the help of this cumulative energy matrix, we find seams with minimum energy in each of the images. Notice that seams found by the algorithm need not be the same on each image (see the images in the last row of figure~\ref{beach}). 

For the given set of LDR images, Let $c_{r}$ and $e_{r}$ denote the minimum energy seam and its energy value in image $I_{r}$ respectively. Now the problem reduces to a decision problem, and the decision is: which seam has to be chosen for insertion or deletion. One option is to take the seam consisting of the least energy, out of these minimum energy seams.

Let $e_k = \min\{e_j : j=1,2,\ldots,N \}$  

In this case, seam $c_{k}$, seam with minimum will be deleted from each of the input images.

One can also choose the seam having energy value which is the median of all minimum energy seams in different images. In either case, the accuracy of results solely depends upon the natural scene statistics. In our experiments, we found that median serves better than the minimum. This is due to the fact that the median represents average exposure value from the given set of LDR images.

It may be noted that for $r \neq k$, seam $c_k$ might not be the seam with minimum energy in the image $I_r$. Thus, by deleting or adding the seam $c_k$ in image $I_r$ we might not add or delete the seam with minimum energy. But because we need to maintain the corresponding coordinates, the same seam needs to be added or deleted in all the LDR images. 

We can further improve this strategy by making sure that each time the minimum total energy seam should be added or removed from the final image. As noticed earlier, while removing minimum energy seam $c_{k}$ (which is minimum energy seam for image $I_k$) from the image $I_r$, we might delete a seam with higher energy. To overcome this we remove or add minimum total energy.

Let $s_{ij}$ be the replica seam in image $i$ of the seam with minimum energy in image $j$. If seam $c_{j}$ is deleted from each of the input images, the total energy added or removed is: 

\begin{eqnarray}
&& \hskip .25\columnwidth E_{j}=\sum_{i=1}^N{ \phi (s_{ij})},  1\leq j\leq N \\ 
&& \text{where }, \phi(s_{ij})\text{denotes the energy of the seam} s_{ij} \nonumber \\
&& \hskip 0cm \nonumber \\
&& \hskip .25\columnwidth E_{k}=\min\{E_{j} :   1\leq j\leq N \}  \nonumber 
\end{eqnarray}

In this case the total amount of energy removed or added will be $E_k$ and desired seam will be $c_{k}$. With this approach we get better results. figure~\ref{chameleon}(c) shows the results after applying this approach.

However while adopting the statistical approach discussed, the seam having the least energy need not be one among the candidate low energy seams. This does not guarantee the removal or the addition of the desired least energy seam. This is due to the fact that while calculating the total minimum energy we are only concerned about the energy of the candidate low energy seams in each image. Other possible seams which could have lead to a much better solution to the problem we address are discarded. Therefore this approach is not the optimal one. 

\subsection{Aggregate Energy Metric Approach}

Instead of finding energy matrices for each of the input LDR images separately, we can think of an aggregate energy matrix. Now we generate a aggregate cumulative energy matrix from this aggregate energy matrix. This aggregate cumulative energy matrix should be generated in such a way that any seam which is indicated as minimum energy seam by this matrix should be of least importance. This criterion is necessary because it guarantees that we will not lose important information during retargating.

For example, if we are taking magnitude of gradient as our energy metric (For individual LDR images) then our aggregate energy metric will be a function of gradients of individual images.
\begin{equation}
\hskip .25\columnwidth E=f(E_{1},E_{2},E_{3},...E_{N})
\end{equation}

In this work we have defined this function as a linear combination of the gradient of each LDR image.

\begin{eqnarray}
&& \hskip .25\columnwidth E=\sum_{i=1}^{N}{\alpha_{i}E_{i}}\\
&& \hskip .2\columnwidth \text{where}, \ \ \displaystyle\sum_{i=1}^{N}{\alpha_{i}}=1 \nonumber
\end{eqnarray}
Parameter $\alpha_{i}$ corresponds to weight given to image $i$ in aggregate energy metric. Now through this aggregate energy metric $E$ our algorithm generates an aggregate cumulative energy metric which defines the energy level for seams. Weight parameter $\alpha_{i}$ should be chosen in such a way that region which are underexposed or overexposed in the LDR images will get lesser weight compared to other regions. 

Average energy per pixel in each image could be used as a weighting parameter. 
$\alpha_{i}$ is the average energy per pixel in the $i^{th}$ image.

This weighting parameter has an important role in making the decision regarding which seam needs to be added or deleted. We further try to calculate this weighting parameter using some other image characteristics. Laplacian of an image calculates second derivative along both the spatial directions (horizontal as well as vertical) and this Laplacian indicates sharp edges in the image. Therefore it will serve better for calculating the weight parameter.

We calculate weighted Laplacian for each image and then perform an element wise multiplication of this with the energy matrix of each image and then calculate the summation, this matrix will now work as the aggregate energy metric.

\begin{eqnarray}
&& \hskip .25\columnwidth E(x,y)=\sum_{i=1}^{N}{L_{i}^{*}(x,y)*E_{i}(x,y)}\\ 
&& \hskip .2\columnwidth \text{where}, \ \ L_{i}^{*}(x,y)={L_{i}(x,y) \over \sum_{i=1}^{N}{L_{i}(x,y)}}  \nonumber 
\end{eqnarray}

Here both multiplication and division are performed element wise.

With this approach (Aggregate energy metric with weighted Laplacian as a weighting parameter), the final image is not only losing (or adding in case of enlarging images) minimum energy but also the output resized HDR image will be of better quality than the direct approach.
\begin{figure}[!htbp]
\centering
\subfigure[]{\includegraphics[width=.3\columnwidth]{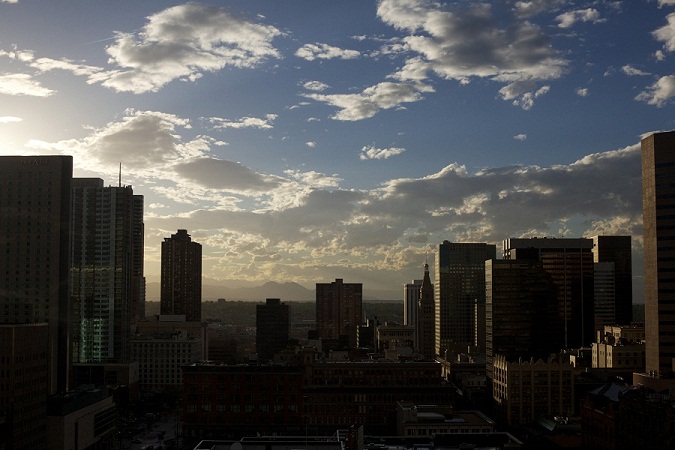}
\includegraphics[width=.3\columnwidth]{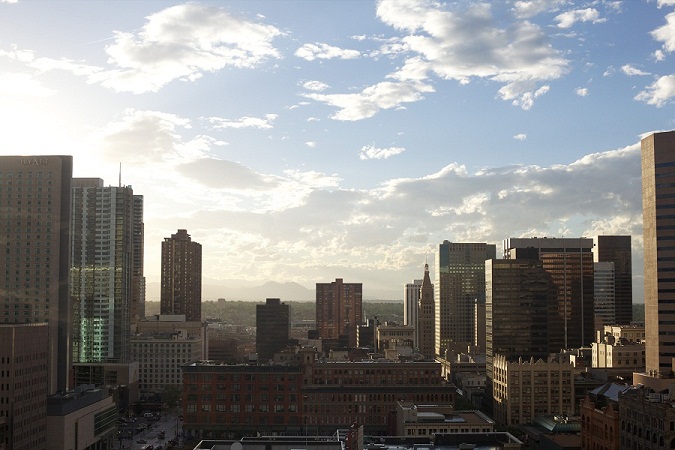}
\includegraphics[width=.3\columnwidth]{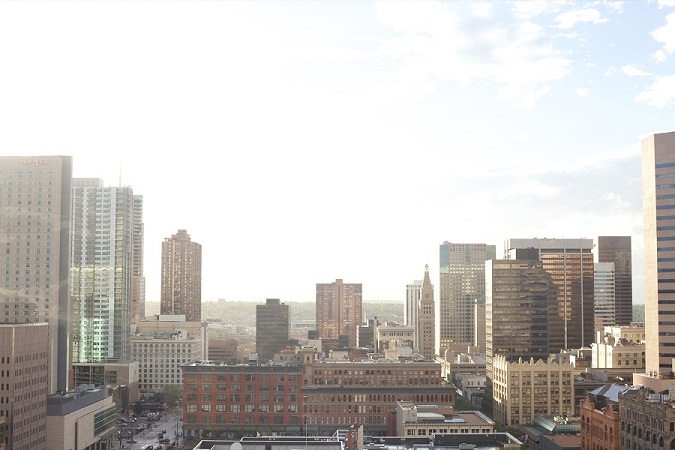}}

\subfigure[]{\includegraphics[width=.4\columnwidth]{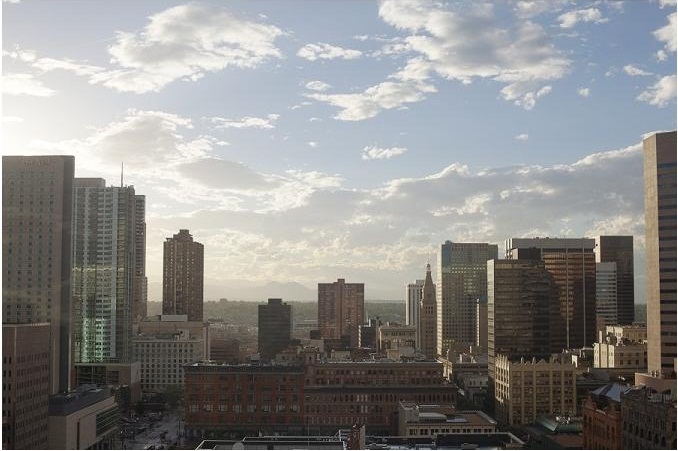}
\includegraphics[width=.4\columnwidth]{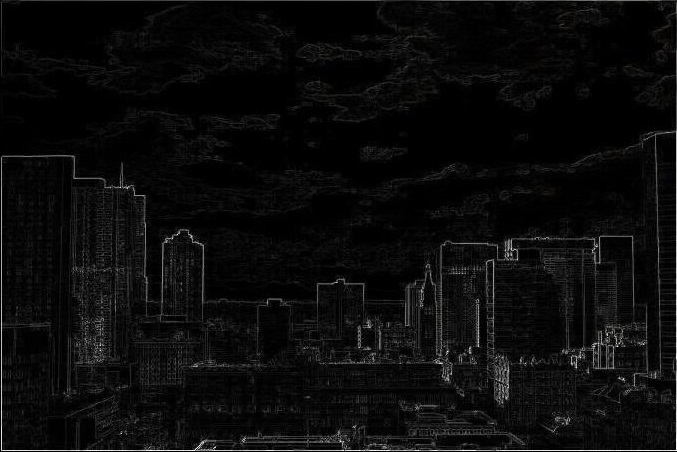}}

\subfigure[]{\includegraphics[width=.3\columnwidth]{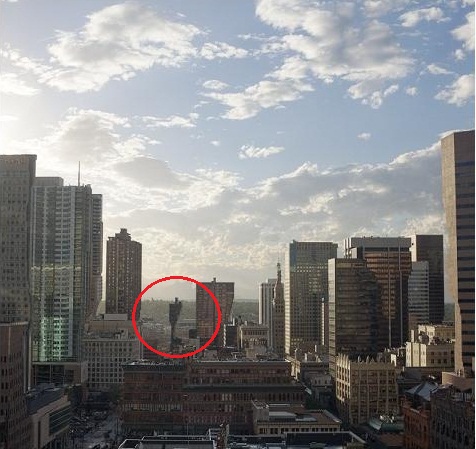}}
\subfigure[]{\includegraphics[width=.3\columnwidth]{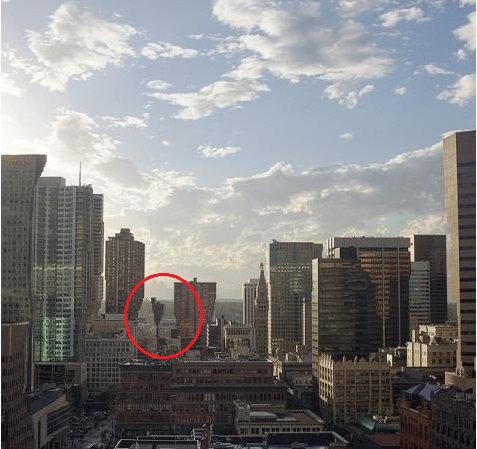}}
\subfigure[]{\includegraphics[width=.3\columnwidth]{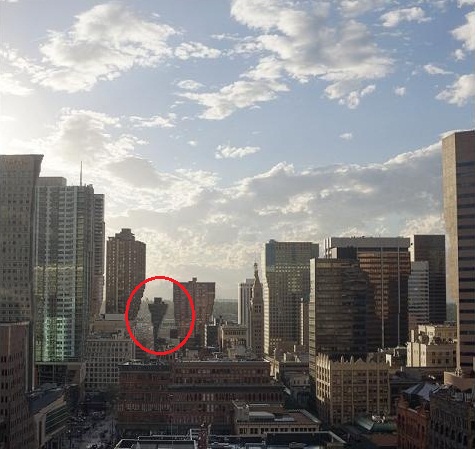}}

\caption{LDR image results when reducing the aspect ratio of the image (30\%). (a) Exposure sequence, (b) Exposure fused image and its energy distribution, (c) Direct approach, (d) Statistical Approach. (e) Aggregate Energy Metric approach (weighted using average energy). Marked region shows how different approaches affect the content in that particular region of the scene in the LDR images.}
\label{house}
\end{figure}

\begin{figure}[!htbp]
\centering
\subfigure[]{\includegraphics[width=.8\columnwidth]{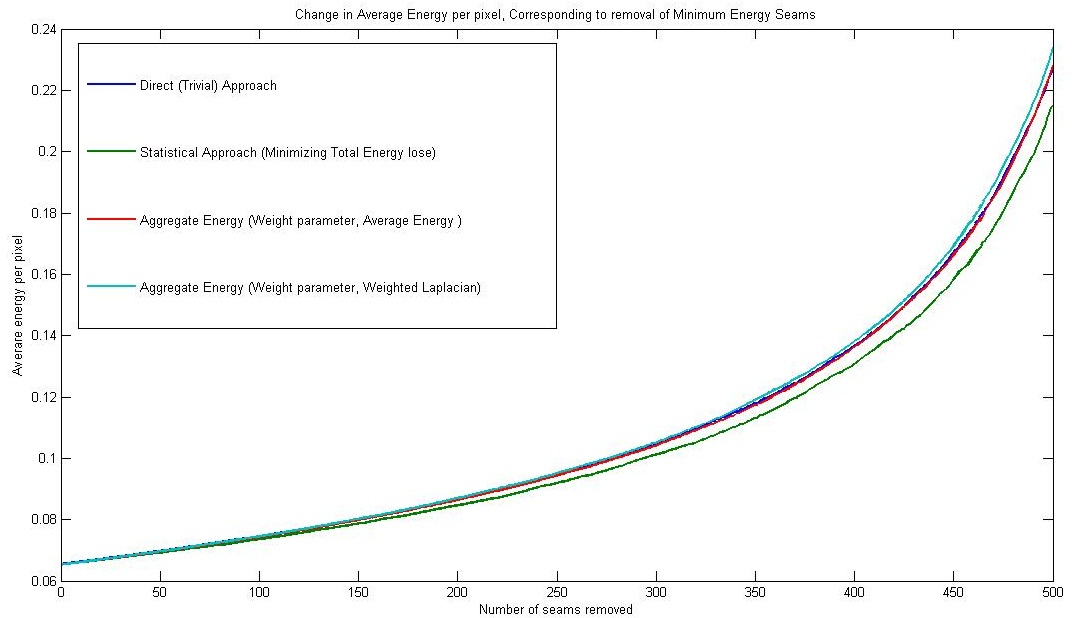}}
\subfigure[]{\includegraphics[width=.8\columnwidth]{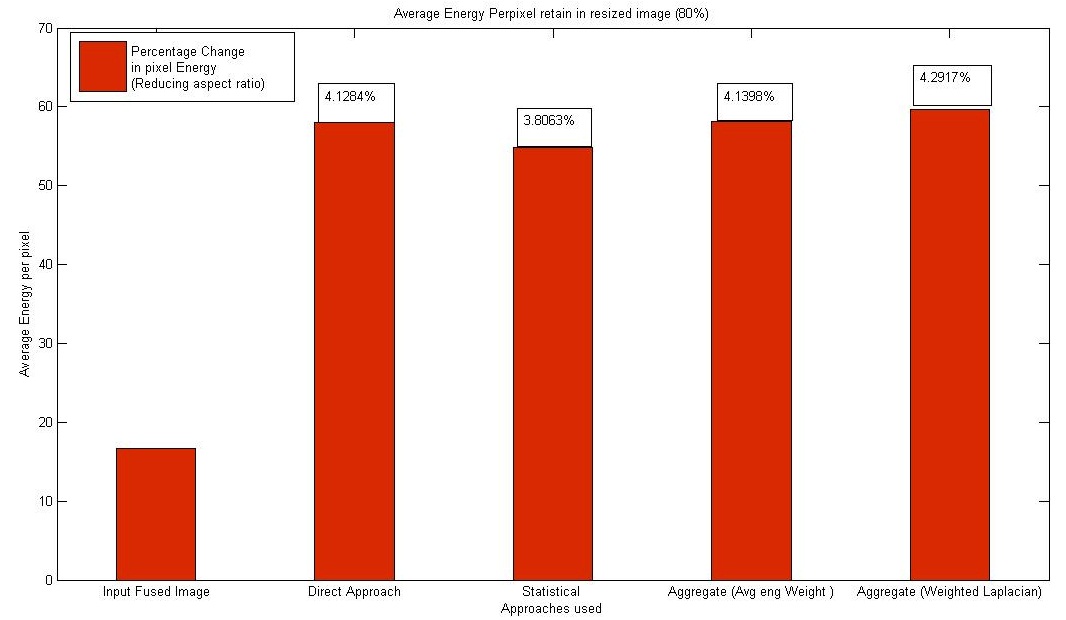}}

\caption{Change in average energy per pixel in final LDR images generated using various approaches. (a) Comparison between the total energy in final output high Contrast LDR image through these approaches, Plot shows that upto certain limit the proposed approach is similar to that of direct approach. But after that, the proposed approach preserves more energy. (b) Average energy of input image and resized image using different approaches.}
\label{plots}
\end{figure}

\section{Results}
\label{result}

\begin{figure*}[!htbp]
\centering
\subfigure[]{\includegraphics[width=.2\textwidth]{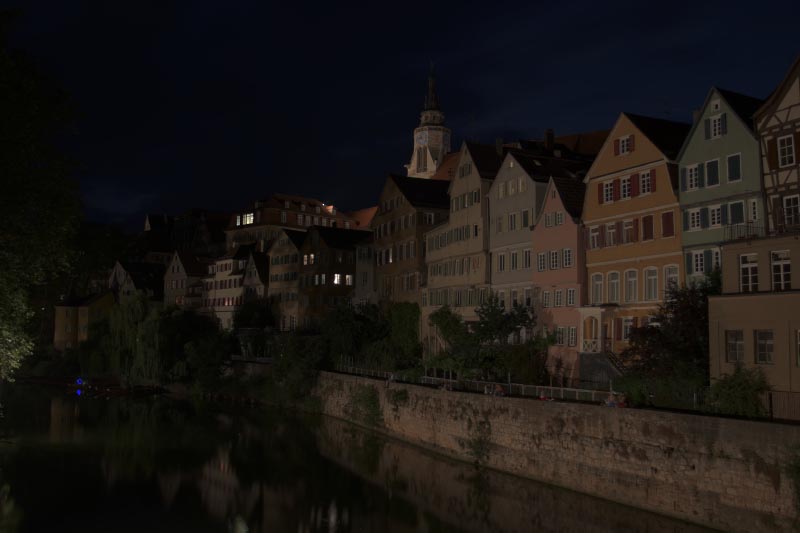}
\includegraphics[width=.2\textwidth]{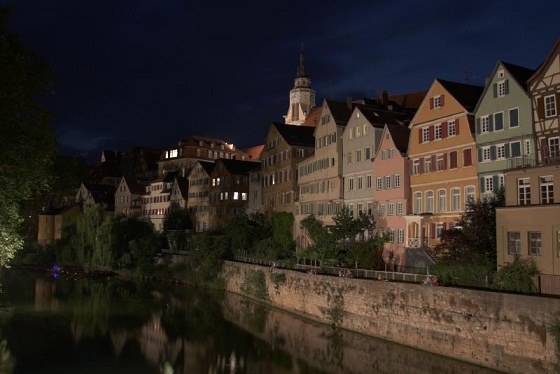}
\includegraphics[width=.2\textwidth]{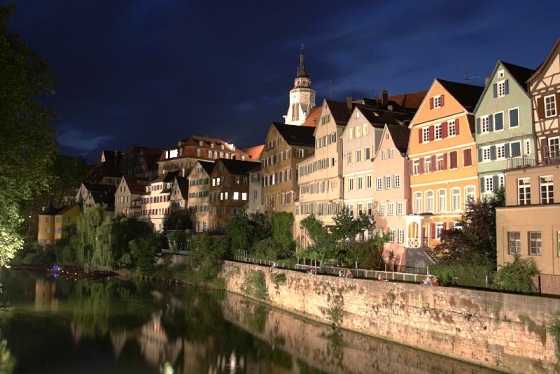}
\includegraphics[width=.2\textwidth]{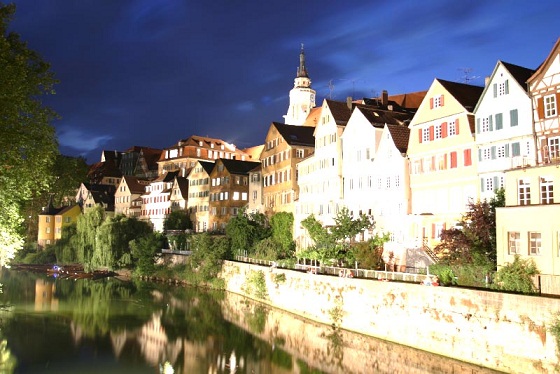}
\includegraphics[width=.2\textwidth]{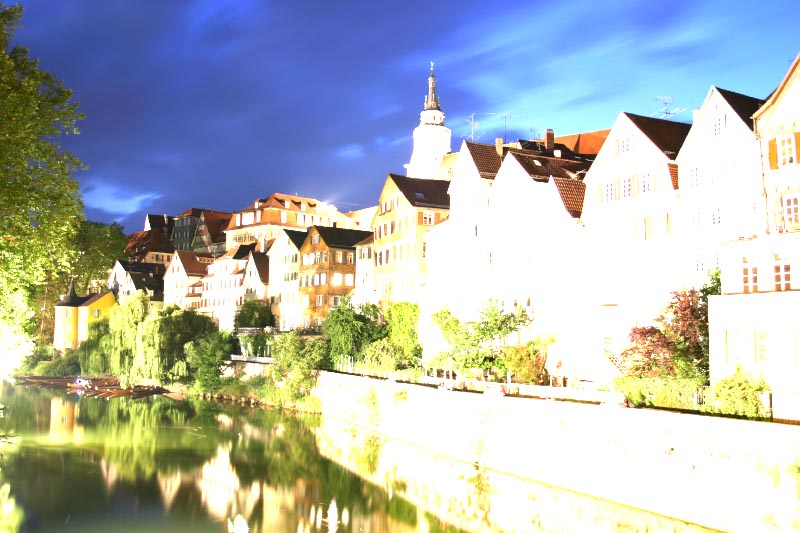}}

\subfigure[]{\includegraphics[width=.16\textwidth]{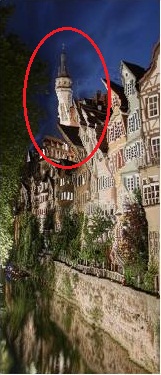}\includegraphics[width=.16\textwidth]{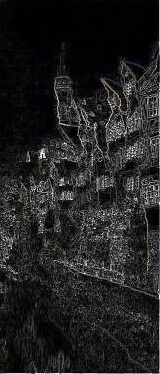}}
\subfigure[]{\includegraphics[width=.16\textwidth]{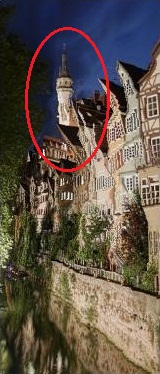}\includegraphics[width=.16\textwidth]{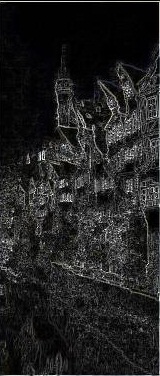}}
\subfigure[]{\includegraphics[width=.16\textwidth]{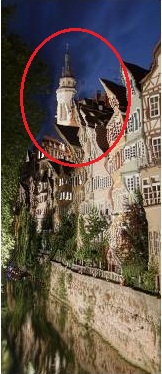}
\includegraphics[width=.16\textwidth]{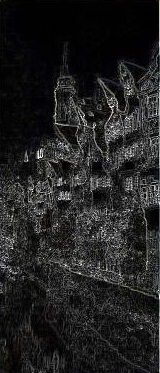}}

\caption{Comparison between the energy distribution of final resized high contrast LDR images created through various approaches. Images are reduced horizontally to 70\% of their original resolution. (a) Input exposure time sequence, (b) Direct approach, (c) Statistical approach (minimization of total minimum energy), and (d) Aggregate energy metric approach with weight assigned according to average pixel energy, (b-d) Left: resized LDR image, Right: Energy Distribution of resized LDR images. Marked region shows visually how the proposed approach works better in preserving content information in that region. }
\label{tamp}
\end{figure*}

In this section we presents results achieved by our algorithm using various approaches discussed above. Our main concern is not to lose (or add in case of enlarging) too much energy while resizing. In other words, we want resizing in a content aware manner.
\begin{figure}[!htbp]
\centering
\subfigure[]{
\includegraphics[width=.25\columnwidth]{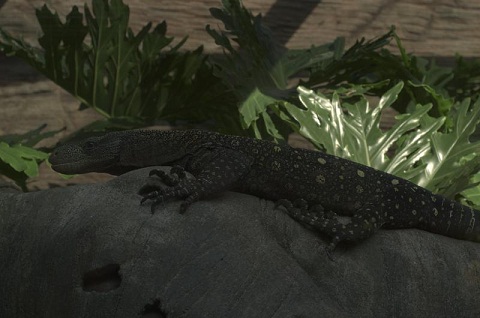}
\includegraphics[width=.25\columnwidth]{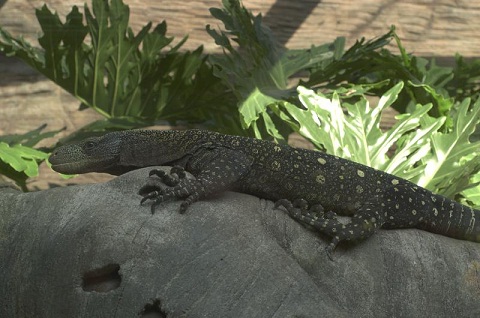}
\includegraphics[width=.25\columnwidth]{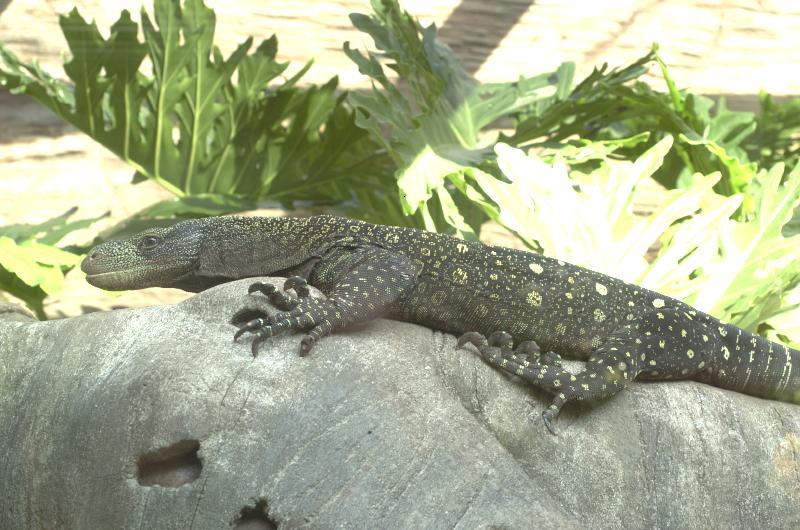}
\includegraphics[width=.25\columnwidth]{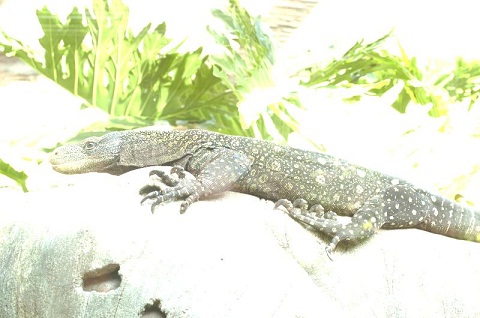}}

\subfigure[]{\includegraphics[width=.48\columnwidth]{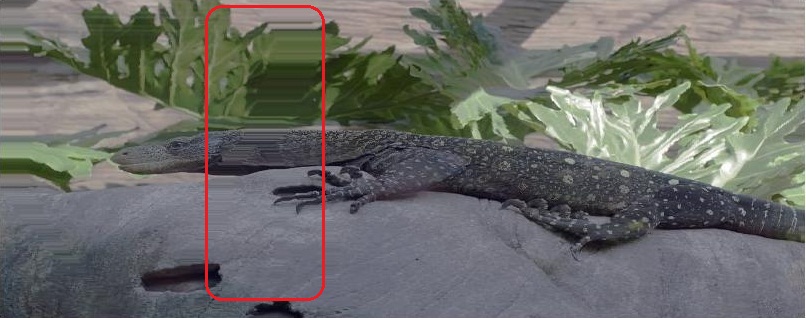}}
\subfigure[]{\includegraphics[width=.48\columnwidth]{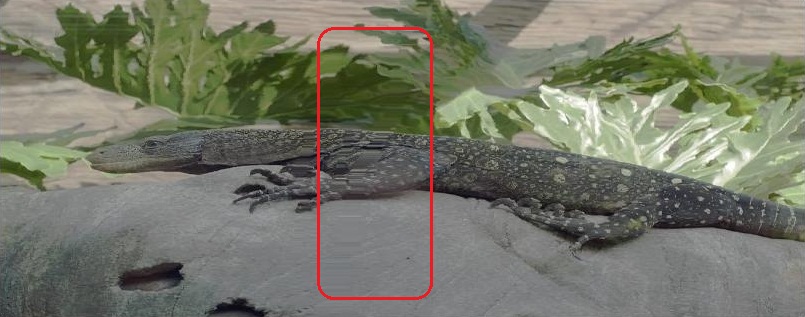}}
\subfigure[]{\includegraphics[width=.6\columnwidth]{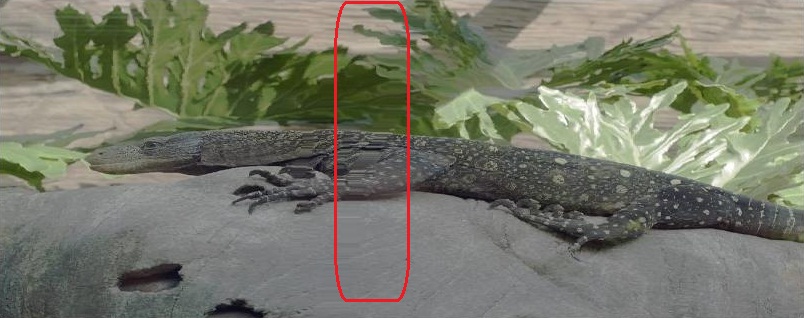}}

\caption{LDR image results while increasing the aspect ratio of output high contrast LDR image(70\% resizing).(a) Exposure sequence, (b)Direct approach, (c) Statistical approach, (d) Aggregate energy metric approach. Marked region indicates that the results of aggregate energy metric weighted using both weighted Laplacian and Average energy per pixel, have better content preservation compared to direct and statistical approaches. Images Courtesy: Erik Reinhard, University of Bristol.}
\label{chameleon}
\end{figure}
Figure~\ref{house}  and figure~\ref{tamp} show the results obtained while reducing the final high contrast LDR image horizontally through various approaches. The marked region indicates how the shape of marked object is affected differently by these methods. One can notice easily that both the improved and the aggregate energy metric approaches preserve the indicated region better than the direct approach.

Figure~\ref{plots}(a) shows change in average energy per pixel, with removal of minimum energy seams over all the different approaches we have discussed. Plot shows that initially all the all the approaches works similar, but as we move to the higher degree of resizing (in this case compression) the behavior of various approaches changes. Plot clearly shows that aggregate energy metric with weighted Laplacian as a weighting parameter will preserve the highest energy. Figure~\ref{plots}(b) show the quantitative information about how much energy is preserve through various approaches.

\begin{figure}[!htbp]
\centering

\subfigure[]{\includegraphics[width=.45\columnwidth]{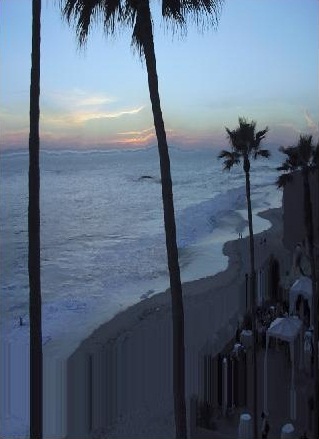}\includegraphics[width=.45\columnwidth]{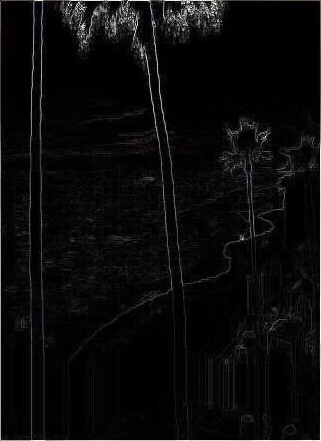}}

\subfigure[]{\includegraphics[width=.45\columnwidth]{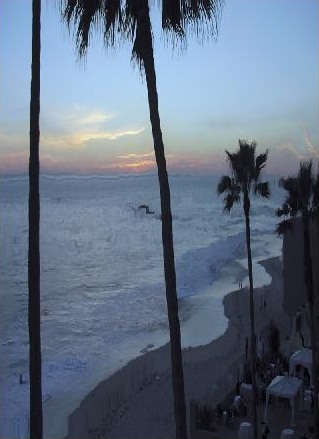}
\includegraphics[width=.45\columnwidth]{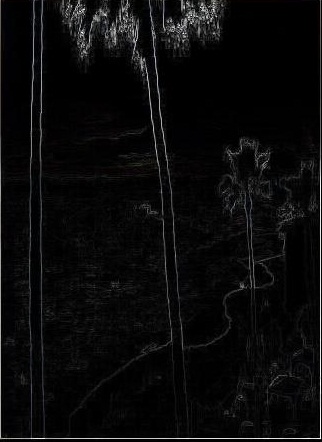}}

\caption{Image resizing in height and corresponding energy distribution of exposure sequence (see figure~\ref{beach}). (a) Direct approach, (b) Aggregate energy metric approach with weighted Laplacian as the weighting parameter.}
\label{beach1}
\end{figure}

Figure~\ref{chameleon} shows the results while enlarging the final high contrast LDR image by inserting seams through various techniques. It can be seen that aggregate energy metric approach (figure~\ref{chameleon}(d)) yields the best result.  Figure~\ref{beach1}(a) shows how the artifacts are introduced while enlarging the input images by the direct approach beyond a certain limit. However, in the same case (see figure~\ref{beach1}(b)) aggregate energy metric with Laplacian as weight parameter yields good results. The respective energy distributions are shown alongside.

\section{Conclusion}
\label{con}
We have proposed novel approach for the content aware resizing of multi-exposure images of a static HDR scene before fusing them into a high contrast LDR image. The proposed approach efficiently combines the content aware image retargeting and the multi-exposure images to develop a novel application suitable for any digital device. We showed that the proposed algorithm performs better when compared to the direct approach of fusing the multi-exposure images before content aware resizing. We have shown through experiments that the LDR image results generated using the proposed statistical and aggregate energy metric approaches to be far better both visually as well as energy preserving criteria. The optimal selection of seams to insert or delete leads to highly robust retargeting algorithm. The proposed approach is fully automatic with no user intervention. The proposed algorithms open up a wide possibility of retargeting and fusion techniques which can be customized for a given display device. 

As the approach does not involve any iterative solution or minimization of any complex cost function, it is computationally inexpensive. The developed algorithms can either be included along with the state of the art mobile cameras/digital cameras and can be provided as applications for post capture image processing softwares. The proposed approaches assume perfectly registered images of a static scene which is a hard constraint to be placed on a real world scene. We hope that the proposed approach can be improved and extended in the case of dynamic scenes which tend to introduce ghosting artifacts. Further, we hope to extend this approach for video image retargeting applications involving HDR scenes. We believe that the novel approach discussed here would lead to more novel ideas in the flexible resolution image retargeting research.

\bibliographystyle{IEEEtran}
\bibliography{ref}

\end{document}